\documentclass[11pt]{article}

\usepackage[final]{acl}

\usepackage{times}
\usepackage{latexsym}
\usepackage{mathtools}
\usepackage{amssymb}
\usepackage{bm} 
\usepackage{tabularx}
\usepackage{booktabs}

\usepackage[T1]{fontenc}

\usepackage[utf8]{inputenc}

\usepackage{microtype}

\usepackage{inconsolata}

\usepackage{graphicx}

\title{Inferring L1 Backgrounds from L2 English with LLM Token Probabilities}

\author{
  \textbf{Ahmet Yavuz Uluslu}$^{1,2, 3}$ \quad
  \textbf{Mark Gales}$^1$ \quad
  \textbf{Kate Knill}$^1$ \quad
  \textbf{Gerold Schneider}$^{2}$ \\[0.2cm]
  $^1$ALTA Institute, Department of Engineering, University of Cambridge \\
  $^2$Department of Computational Linguistics, University of Zurich 
  \\
  $^3$PRODAFT, Yverdon-les-Bains, Switzerland \\
  \tt{\{ayu20, mjfg100, kmk1001\}@cam.ac.uk} \\
}

\begin{document}
\maketitle
\begin{abstract}
Native language identification (NLI) infers a non-native speaker’s first
language (L1) from their second language (L2) production. We investigate a token probability approach that scores each text sample as a continuation of a prompt prefix specifying a candidate L1, then predicts the candidate with the highest score. We evaluate our approach using Llama 3.1 8B and 70B on three datasets: TOEFL9, Write \& Improve 2024, and BULATS-L1. We compare its performance with previous prompting-based classification approaches. Token-probability scoring substantially improves on direct classification, particularly with the 8B model, while the 70B model achieves state-of-the-art performance across all datasets. Our results indicate that pretrained LLMs’ token probabilities contain information useful for distinguishing L1 backgrounds from L2 English production.
\end{abstract}

\section{Introduction}

Native Language Identification (NLI) is the task of inferring a non-native speaker's first language (L1) from their production in a second language (L2). NLI is motivated in part by cross-linguistic influence: a learner's L1 can shape their L2 production and may leave patterns associated with that background \citep{jarvis2008crosslinguistic}. Strong performance by large language models (LLMs) on related tasks, such as grammatical error detection and multilingual translation, has motivated interest in their use for NLI \citep{goswami2024native}. Recent studies have explored zero-shot, prompting-based classification approaches to NLI \citep{zhang2023native, ng2025leveraging, goswami2025multilingual}, with some reporting state-of-the-art (SoTA) results on NLI benchmarks. More recent work has also shown that editorial corrections to learner text can change the probability an LLM assigns to its predicted L1 label \citep{uluslu2026impact}. This motivates us to examine the token-level probabilities the model assigns to the essay under different contexts, rather than only its generated answer.

We investigate a token-probability approach in which a pretrained LLM scores an essay under prompts specifying different
candidate L1 backgrounds. The model predicts the candidate whose
prompt gives the essay the highest score. We compare this approach with direct LLM classification and examine
how performance changes with the amount of text available. We evaluate our approach using Llama 3.1 8B Instruct and Llama 3.1 70B Instruct \citep{grattafiori2024llama} on three datasets: TOEFL9 \citep{blanchard2013toefl11}, Write \& Improve 2024 \citep{nicholls2024write}, and BULATS-L1 \citep{anonymous2026acoustic}.

Our findings show that token-probability scoring substantially improves on direct classification, particularly with the 8B model. This is notable given prior findings that models of this scale may require task-specific fine-tuning to achieve competitive performance on NLI \citep{ng2025leveraging}. The 70B model achieves SoTA performance across all three datasets in the evaluation settings considered. Our results show that token probabilities conditioned on candidate-L1 prefixes provide a useful signal for distinguishing L1 backgrounds from L2 English production.

\section{Native Language Identification}
\label{sec:nli}

\paragraph{Classification and chain-of-thought approaches.}
One common approach is to prompt a large language model (LLM) to
classify the essay directly. The model receives natural language instructions, the
essay, and a set of candidate L1 labels, and is asked to return a
label \citep{zhang2023native, ng2025leveraging, goswami2025multilingual}. Related concurrent work by \citet{anonymous2026acoustic} compares speech transcripts and written texts within this L1-classification framework. Another approach asks the model to produce a rationale or
chain-of-thought (CoT) before returning its prediction \citep{zhang2023native}. But a generated rationale does not, by itself, show that the model systematically evaluated the essay under each candidate
background or that the explanation faithfully reflects the basis for
its prediction \citep{uluslu2025robust}. Recent work has shown that correcting grammatical errors and altering
word order can change the probability an LLM assigns to its predicted L1 label \citep{uluslu2026impact}. This motivates us to examine the
probabilities assigned to essay tokens under different candidate-L1
contexts.

\paragraph{Language-modeling for learner language.}
The use of model surprisal and essay-level probabilities for this task
has also been explored in earlier work
\citep{lotfi2020deep}. These approaches train
language models such as GPT-2 on learner essays associated with
particular L1 backgrounds and compare how well each model predicts a
test essay. For a model $M_\ell$ associated with background $\ell$,
the average token-level negative log-likelihood of an essay
${\bm x}_{1:T}=(x_1,\ldots,x_T)$ is
\begin{equation}
  \mathcal{L}_{M_\ell}({\bm x}_{1:T})
  =
  -\frac{1}{T}
  \sum_{t=1}^{T}
  \log p_{M_\ell}(x_t \mid {\bm x}_{<t}).
  \label{eq:background-lm-loss}
\end{equation}
For NLI, the essay is assigned to the background whose model yields the
lowest loss, or equivalently the highest length-normalized
log-probability.

\paragraph{Scoring the essay under candidate prefixes.}
Our approach differs in how the candidate L1 information enters the
model. Rather than training a separate language model on essays from
each background, we use a shared, general-purpose LLM and condition it on the candidate L1 through the prompt. This lets us investigate whether a model contains useful information for detecting cross-linguistic influence in L2 English writing \citep{acharya2025tracing}. We provide a candidate L1 in the prompt and score the
probability of the observed essay as a continuation. For each
$\ell \in \mathcal{L}$, we use the same prompt template, for example:
\begin{quote}
\small
An English learner whose first language is \texttt{\{L1\}} wrote the
following text:

\texttt{\{learner\_text\}}

\end{quote}
Let $c_\ell$ denote this prompt instantiated with candidate background
$\ell$. Let $p_M$ denote the next-token distribution of the shared model $M$. We define the candidate-conditioned essay score as
\begin{equation}
  S_M(\ell,{\bm x}_{1:T})
  =
  \sum_{t=1}^{T}
  \log p_M\!\left(x_t \mid c_\ell,{\bm x}_{<t}\right).
  \label{eq:essay-score}
\end{equation}
In computing the score, we include the
essay tokens but not the prompt tokens. The model-based prediction is the candidate under which the observed
essay receives the highest score:
\begin{equation}
  \hat{\ell}
  =
  \arg\max_{\ell \in \mathcal{L}} S_M(\ell,{\bm x}_{1:T}).
  \label{eq:likelihood-classification}
\end{equation}
Thus, every L1 candidate is evaluated against the same essay, with the
candidate L1 supplied as prefix. 

\paragraph{Relation to forensic likelihoods.}
Our approach connects to forensic text comparison
\citep{ishihara2024validation} because both assess evidence relative to
competing hypotheses. Let $H_\ell$ be the hypothesis that the writer's
L1 is $\ell$, and the evidence be the observed essay, ${\bm x}_{1:T}$. A forensic
comparison between two hypotheses is commonly expressed as a
log-likelihood ratio:
\begin{eqnarray}
  \operatorname{LLR}_{a,b}(E)
  \!\!\!\!&=&\!\!\!\!
  \log\left(\frac{p({\bm x}_{1:T} \mid H_{\ell_a})}
       {p({\bm x}_{1:T} \mid H_{\ell_b})}
       \right) \nonumber\\
  \!\!\!\!&\approx &\!\!\!\!   S_M(\ell_a,{\bm x}_{1:T}) - S_M(\ell_b,{\bm x}_{1:T}).
  \label{eq:forensic-lr} \hspace{1.01em}
\end{eqnarray}
In our approach, the candidate-L1 prefix conditions the model on each hypothesis before observing the evidence.

\section{Datasets}

\subsection{TOEFL9}
We use the test set of TOEFL11, a standard benchmark for NLI \citep{blanchard2013toefl11}, and retain the following nine L1 backgrounds: Arabic, Chinese, French, German, Hindi, Italian, Japanese, Spanish, and Turkish. We exclude Korean and Telugu because they are not represented in either of the other two datasets used in this study. We refer to this nine-language subset as TOEFL9. It contains 900 essays, with 100 essays per L1. The mean essay length is   348 words.

\begin{figure*}[t]
  \centering
  \includegraphics[width=\textwidth]{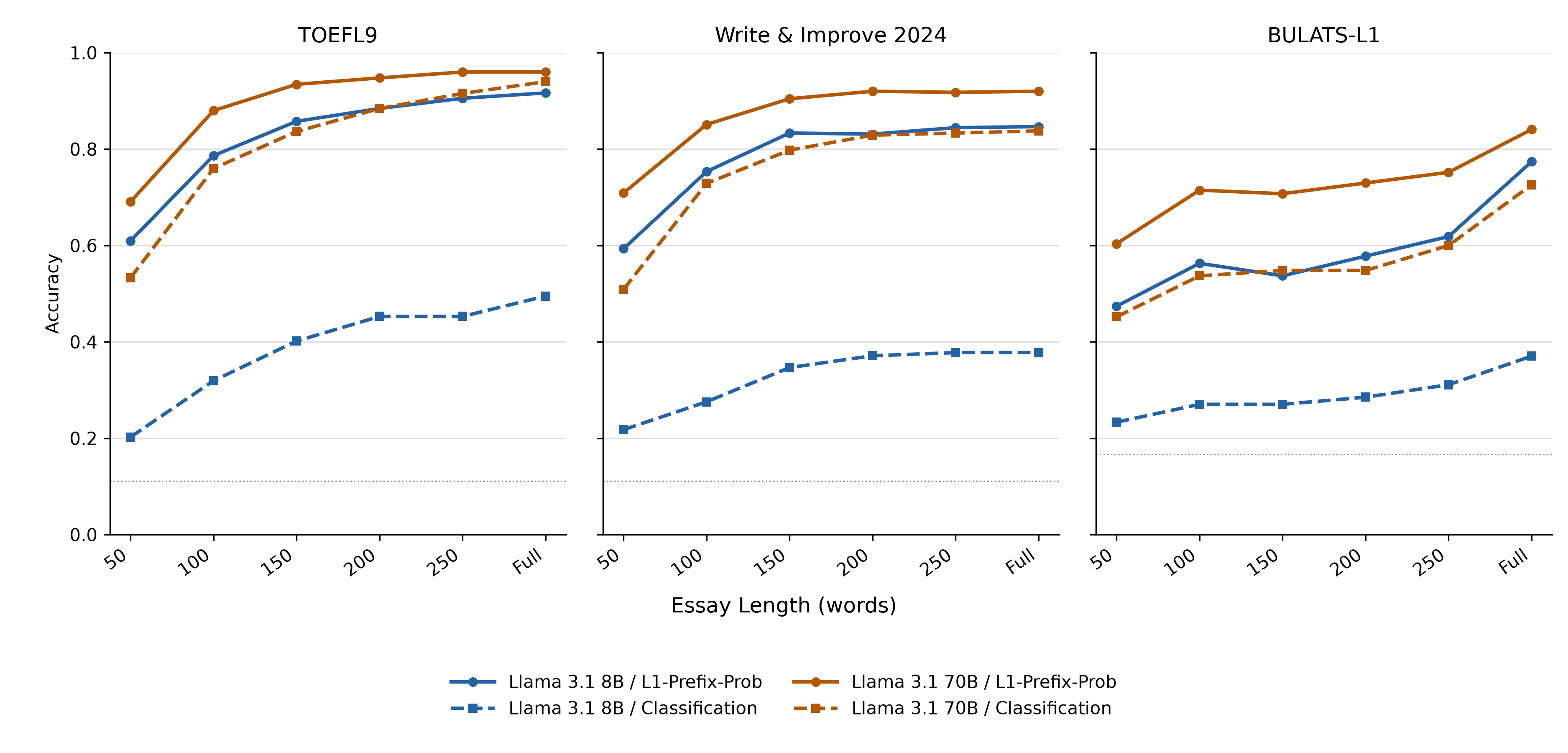}
  \caption{Accuracy as a function of input length for Llama 3.1 8B and
  70B using language-prefix likelihood scoring (solid lines) and direct
  classification (dashed lines). Panels show results on Write \&
  Improve 2024, TOEFL9, and BULATS-L1.}
  \label{fig:length-model-method-comparison}
\end{figure*}

\subsection{Write \& Improve 2024}
\label{sec:wi2024}

We use a subset of the Write \& Improve (W\&I) 2024 corpus
\citep{nicholls2024write} selected for NLI evaluation by
\citet{anonymous2026acoustic}. It contains 450 essays from nine L1
backgrounds, matching those in TOEFL9: Arabic, Chinese, French,
German, Hindi, Italian, Japanese, Spanish, and Turkish, with 50 essays
per L1. The subset includes essays of at least 100 words that are
annotated at CEFR level A2 or above. The mean essay length is
$193 \pm 61$ words.

\subsection{BULATS-L1}
We evaluate our method on BULATS-L1 \citep{anonymous2026acoustic}, a dataset derived from the
Business Language Testing Service (BULATS), developed by Cambridge
English. It contains spoken English submissions from learners with six
self-reported L1 backgrounds: Arabic, Chinese, French, Italian,
Japanese, and Spanish. We use the balanced test set, which comprises
270 submissions, with nine submissions for each L1--CEFR-level
combination across levels A1--C1. The dataset is provided with
automatic speech recognition (ASR) transcripts of the spoken
responses. The mean transcript length is 472 words per submission.

\section{Experiments}

\subsection{Models and Prediction Approaches}

We evaluate two commonly used open-weight language models for NLI: Llama 3.1 8B
Instruct\footnote{\url{https://huggingface.co/meta-llama/Llama-3.1-8B-Instruct}}
and Llama 3.1 70B
Instruct\footnote{\url{https://huggingface.co/meta-llama/Llama-3.1-70B-Instruct}} \citep{ng2025leveraging, nicholls2025cross}. We evaluate each model using two approaches. For direct
classification, we prompt the model to select the writer's L1 from
the candidate languages. The classification prompt is provided in
Appendix~\ref{app:classification-prompt}. For token probability scoring, we follow
Equation~\ref{eq:essay-score}: for each candidate L1, we instantiate
the corresponding prefix and sum the log probabilities assigned to the
input-text tokens. The prefix prompts are provided in
Appendix~\ref{app:prefix-prompt}. For TOEFL9 and Write \& Improve, we
use the written text prefix template. For BULATS-L1, initial
experiments with this template did not produce satisfactory results.
We therefore use a speech-aware prefix that identifies the input as a
transcript of an English learner's spoken response. We predict the
candidate L1 whose prefix yields the highest score. Before inference,
we mask named entities, including locations, nationalities, and
personal names, as well as non-English words. This approach aims to
reduce reliance on explicit identity cues and topic-related shortcuts
\citep{uluslu2025robust}. We compare our methods with state-of-the-art NLI approaches reported in prior work \citep{zhang2023native, uluslu2026impact,
anonymous2026acoustic}. All of these approaches use GPT-4o
\citep{hurst2024gpt} as their best performing model. We report comparisons for datasets and
evaluation settings for which results are available. 

\subsection{Data Efficiency}
To examine how performance changes with the amount of text available,
we evaluate each method using the first 50, 100, 150, 200, and 250
words of each text. We also report results using the full, untruncated
text (``Full''). We use the resulting text as input to both the direct-classification and token probability methods.

\section{Results}

\subsection{Comparison of Prediction Methods Across Input Lengths}

Figure~\ref{fig:length-model-method-comparison} shows accuracy as a
function of input length. Across all three datasets, both model sizes,
and every evaluated length, L1-prefix probability scoring outperforms
direct classification. The difference is especially pronounced for
Llama 3.1 8B. With unrestricted text, direct classification achieves
roughly half the accuracy of prefix scoring across the three datasets.

The 70B model achieves higher accuracy than the 8B model when both use
the same prediction method. However, prefix scoring with the 8B model
matches or exceeds direct classification with the 70B model in several
settings, including full-text Write \& Improve and BULATS-L1. This
shows that the choice of prediction method has a substantial effect,
even when comparing models of different sizes.

Accuracy generally increases as more text is provided, with large
gains often occurring between 50 and 150 words. The improvements
become smaller at longer input lengths in some settings, and there are
occasional small decreases at intermediate lengths. Additional text
is particularly beneficial for BULATS-L1 when the unrestricted
transcript is used. Overall, more textual evidence tends to help, though the size of the improvement varies across datasets and methods.

\subsection{Comparison with Previous Results}

\begin{table}[ht!]
\centering
\small
\begin{tabularx}{\linewidth}{@{}lXr@{}}
\toprule
Dataset & Previous SoTA & Our approach \\
\midrule
TOEFL9 & 95.1\% \citep{zhang2023native} & \textbf{96.0\%}\\
W\&I 2024 & 88.9\% \citep{uluslu2026impact} & \textbf{92.0\%} \\
BULATS-L1 & 83.3\% \citep{anonymous2026acoustic} & \textbf{84.1\%}\\
\bottomrule
\end{tabularx}
\caption{Full-text accuracy compared with previously reported results. Our approach uses L1-prefix probability scoring with Llama 3.1 70B.}
\label{tab:fulltext-sota-comparison}
\end{table}

Table~\ref{tab:fulltext-sota-comparison} compares full-text results
from Llama 3.1 70B using L1-prefix probability scoring with the best
results reported in prior work. Our method outperforms the reported results on all three datasets. These comparisons
show that an open-weight 70B model using prefix-conditioned token
probabilities can achieve strong performance relative to the
GPT-4o-based approaches reported in prior work.

The TOEFL results should be interpreted in light of concerns about
possible training-data leakage for this benchmark \citep{goswami2025multilingual}.
However, we observe the same broad pattern on Write \& Improve 2024 and BULATS-L1, datasets that were not publicly available when the Llama models were released. This indicates that the findings are not specific to TOEFL9 or dependent only on exposure to that benchmark. Accuracy on BULATS-L1 is lower than on the two written datasets for the 70B prefix-scoring approach. BULATS-L1 consists of transcripts of spoken responses, which may contain disfluencies and ASR errors.  The datasets also differ in other respects, so their raw accuracies should be compared with caution.

\subsection{Score Decomposition}
To illustrate how candidate-L1 prefixes affect scores within a text, we analyze a single sentence from a Turkish-labeled essay using Llama 3.1 8B. For each candidate L1, we center the sentence and word-level subtoken log-probabilities relative to the nine-candidate mean. Appendix~\ref{app:score-decomposition} details the scoring formulation, and Figure~\ref{fig:turkish-first-sentence-llr} shows the resulting sentence-level ranking and word-level contributions.

\section{Conclusion}
We investigated whether pretrained LLMs can identify L1 backgrounds by
scoring learner texts under prompts that specify candidate L1s. Across
three datasets, two model sizes, and all evaluated input lengths,
L1-prefix probability scoring consistently outperformed direct
classification. The results show that even the 8B model can perform
strongly with this approach, while the 70B model achieved full-text
accuracies of 96.0\% on TOEFL9, 92.0\% on Write \& Improve 2024, and
84.1\% on BULATS-L1. These results exceed the previously reported
scores in the evaluation settings considered. Results on the two recently
released datasets also suggest that the findings are not limited to
TOEFL9. Future work should investigate the linguistic and token-level
signals that drive these scores, including how candidate-L1 prefixes
change the probabilities assigned to particular parts of an essay.
Such analyses could help explain what information the models use to
distinguish L1 backgrounds. Overall, prefix-conditioned token
probabilities provide a promising approach to NLI and a useful
complement to direct LLM-based classification.

\section*{Limitations}
\label{sec:limitations}

\paragraph{Dataset representativeness and low-resource scenarios.} Our results are limited to English L2 texts and the L1 backgrounds
included in the three benchmarks. It remains to be seen whether the
gains from L1-prefix-probability scoring hold for other L2 languages
and for L1 backgrounds not represented in these benchmarks, especially
those that may be underrepresented in the models' pretraining data.

\paragraph{Computational cost.}
Prefix scoring requires evaluating the essay under each candidate-L1
prefix, whereas direct classification reads the essay once and
generates a short label. The essay-token probabilities are scored
without autoregressively generating the essay, and candidate scores
can be batched. Prefix scoring therefore involves more
candidate-conditioned computation, but this does not necessarily
translate into proportionally higher inference latency. We did not
measure runtime or resource use, so a direct efficiency comparison
remains for future work.

\section*{Ethical Considerations}
Our research uses publicly available L2 English learner corpora,
including the pseudonymized Write \& Improve corpus
\citep{nicholls2024write} and TOEFL11
\citep{blanchard2013toefl11}. TOEFL11 contains no direct personal
identifiers. We also use speech transcripts from BULATS-L1
\citep{anonymous2026acoustic}, which was anonymized and made available
for this study but is not publicly released. We acknowledge the broad societal implications of authorship analysis, including potential risks to the security and privacy of individuals \cite{saxena2025responsible}. This work is intended to contribute to
educational research on L2 learner language and research on the transparency and privacy implications of automated text analysis. As detailed in our Limitations (Section \ref{sec:limitations}), we also recognize that our efforts to mitigate bias are not exhaustive, and further research is needed.

\section*{Acknowledgments}
This work was supported by Innosuisse innovation project 103.188 IP-ICT (AUCH). The project also received support from Cambridge University Press \& Assessment, a department of The Chancellor, Masters, and Scholars of the University of Cambridge.


\bibliography{custom}

\clearpage
\newpage

\appendix

\section{Direct-Classification Prompt}
\label{app:classification-prompt}

The following prompt was used for direct classification:

\begin{quote}
\small
\begin{verbatim}
You are a forensic linguistics expert
that reads texts written by non-native
authors in order to identify their native
language. Analyze each text and identify 
the author's native language.
Use clues such as spelling errors, word
choice, syntactic patterns, and
grammatical errors to decide.
You ONLY respond in JSON files. The
expected output from you has to be:
{"native_lang": "<language>"}
\end{verbatim}
\normalsize
\end{quote}

\section{Prefix Prompts}
\label{app:prefix-prompt}

\paragraph{Write \& Improve 2024  and TOEFL9}
\begin{quote}
\small
\begin{verbatim}
An English learner whose first 
language is {language} wrote the
following text:

{text}
\end{verbatim}
\normalsize
\end{quote}

\paragraph{BULATS-L1}
\begin{quote}
\small
\begin{verbatim}
The following is a speech transcript
produced by an English learner whose 
first language is {language}:

{text}

\end{verbatim}
\normalsize
\end{quote}

\section{Compute Budget and Model Runs}
The experiments were inference-only, no model fine-tuning was performed. Reported accelerator runtimes were:
\begin{center}
\footnotesize
\setlength{\tabcolsep}{4pt}
\begin{tabularx}{\linewidth}{@{}X p{0.42\linewidth}@{}}
\toprule
Model and accelerator & Runtime and estimated cost \\
\midrule
Llama 3 80B FP16 on H200 SXM (141 GB VRAM) & 3 h at \$5.29/h = \$15.87 \\
Llama 3.1 8B Instruct on A40 (48 GB VRAM) & 5 h at \$0.59/h = \$2.95 \\
\midrule
\textbf{Total} & \textbf{8 h; \$18.82} \\
\bottomrule
\end{tabularx}
\end{center}

\newpage

\section{Exploratory Analysis of L1-Mean-Centered Log-Likelihoods}
\label{app:score-decomposition}

\begin{figure}[p]
\centering
\includegraphics[width=\textwidth]{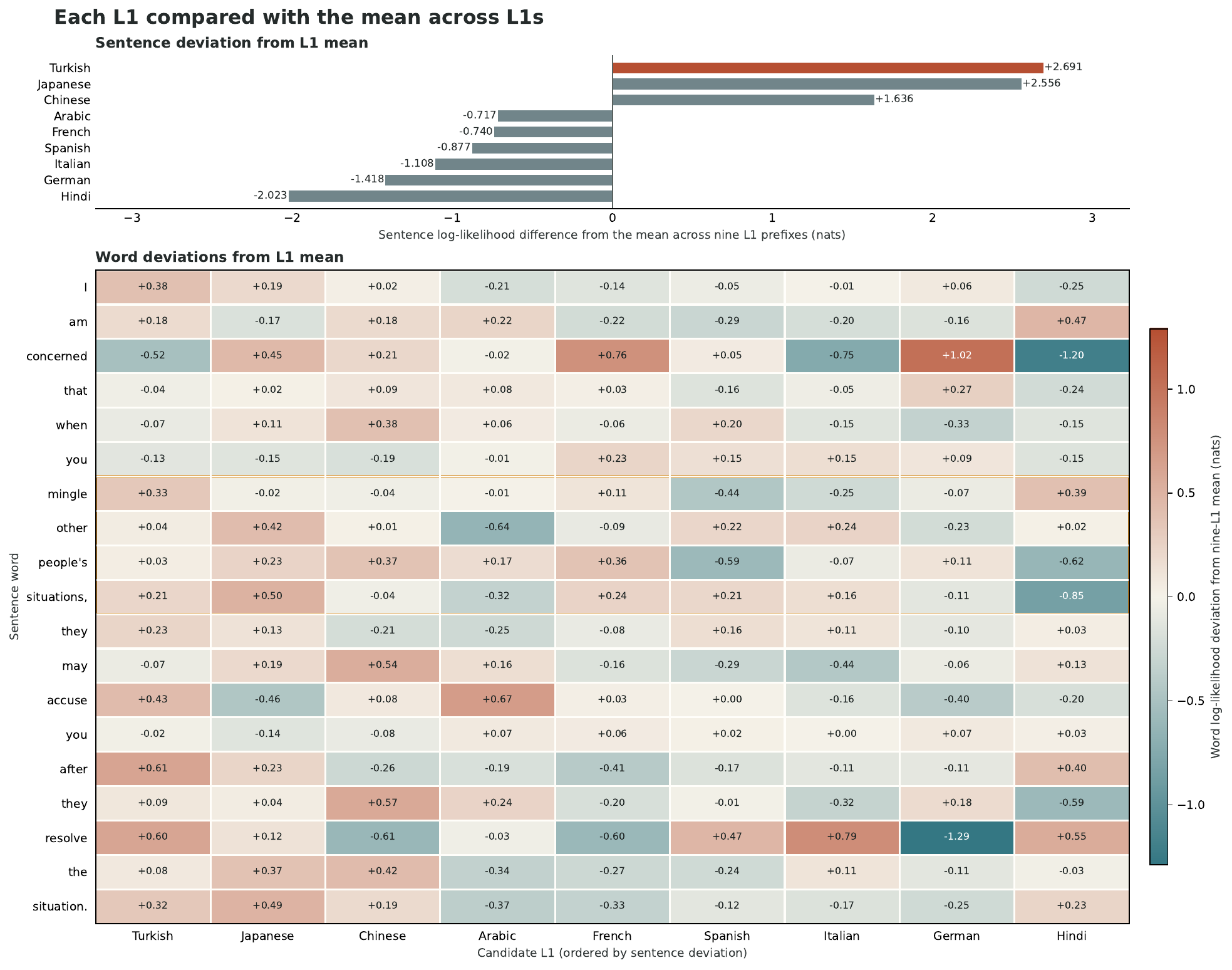}
\caption{Sentence-level ranking and word-level log-probability deviations from the nine-L1 mean (in nats). The upper panel sorts candidate L1s by total sentence deviation ($D_l$), and the lower panel shows word-level deviations ($D_{l,j}$).}
\label{fig:turkish-first-sentence-llr}
\end{figure}

Figure~\ref{fig:turkish-first-sentence-llr} shows the sentence-level score differences and word-level contributions for the nine L1 prefixes.

\subsection{Score decomposition}

For the sentence shown, $T=22$ model subtokens are grouped into 19 whitespace-delimited words in the heat map. Using the candidate-conditioned score $S_M(\ell,\bm{x}_{1:T})$ defined in Equation~\ref{eq:forensic-lr}, the upper panel plots each L1's deviation from the arithmetic mean across the nine candidates:
\begin{equation}
D_\ell=S_M(\ell,\bm{x}_{1:T})-\frac{1}{|\mathcal{L}|}\sum_{k\in\mathcal{L}}S_M(k,\bm{x}_{1:T}),\qquad
\label{eq:sentence-centered-score}
\end{equation}
This is the final sentence score shown, in nats per sentence. Centering preserves the ranking of $S_M$.

For the heat map, each model subtoken is assigned to the whitespace-delimited word with the greatest non-whitespace character overlap, whitespace-only subtokens go to the preceding word, or the first word at sentence start. For the subtokens $A_j$ assigned to word $w_j$, the plotted value is
\begin{equation}
\begin{aligned}
G_M(\ell,w_j) &= \sum_{t\in A_j}
\log p_M\!\left(x_t\mid c_\ell,\bm{x}_{<t}\right), \\
D_{\ell,j} &= G_M(\ell,w_j)-\frac{1}{|\mathcal{L}|}\sum_{k\in\mathcal{L}}G_M(k,w_j).
\end{aligned}
\label{eq:word-centered-score}
\end{equation}
Each heat-map value is a sum of subtoken log probabilities centered on the L1 mean. Positive values mean higher likelihood (lower surprisal) than the mean. Summing word deviations gives $D_\ell$. For example, Turkish is about $+0.33$ nats above the L1 mean for the unidiomatic usage of the word \emph{mingle}.

\onecolumn

\end{document}